\documentclass[12pt]{article}
\usepackage{graphicx}
\usepackage{lipsum}
\usepackage{float}
\usepackage{amsmath}
\usepackage[backend=biber, style=numeric, sorting=none]{biblatex}
\usepackage{url}
\usepackage{doi}
\usepackage{array}
\usepackage{multirow}
\usepackage{hyperref}
\usepackage{caption}
\usepackage{geometry}
\usepackage{amsmath}
\usepackage{booktabs}
\usepackage{tabularx}
\usepackage{rotating}
\usepackage{subfig}
\usepackage{orcidlink}
\usepackage[affil-it]{authblk}
\usepackage{ragged2e}
\usepackage{makecell} 

\title{Enhanced Liver Tumor Detection in CT Images Using 3D U-Net and Bat Algorithm for Hyperparameter Optimization}

\author[1]{Nastaran Ghorbani\thanks{Corresponding author: \href{mailto:nghorbani@luc.edu}{nghorbani@luc.edu} \orcidlink{0009-0008-2262-4101}}}

\author[2]{Bitasadat Jamshidi\thanks{\href{mailto:bitasadat.jamshidi@srbiau.ac.ir}{bitasadat.jamshidi@srbiau.ac.ir} \orcidlink{0000-0002-1397-9065}}}

\author[3]{Mohsen Rostamy-Malkhalifeh\thanks{\href{mailto:mohsen_rostamy@yahoo.com}{mohsen\_rostamy@yahoo.com} \orcidlink{0000-0001-6105-7674}}}

\affil[1]{Department of Mathematics and Statistics, Loyola University Chicago, Chicago, USA}
\affil[2]{Department of Mathematics and Computer Science, Islamic Azad University, Science and Research Branch, Tehran, Iran}

\date{}

\begin{document}
\maketitle

\begin{abstract}
Liver cancer is one of the most prevalent and lethal forms of cancer, making early detection crucial for effective treatment. This paper introduces a novel approach for automated liver tumor segmentation in computed tomography (CT) images by integrating a 3D U-Net architecture with the Bat Algorithm for hyperparameter optimization. The method enhances segmentation accuracy and robustness by intelligently optimizing key parameters like the learning rate and batch size. Evaluated on a publicly available dataset, our model demonstrates a strong ability to balance precision and recall, with a high F1-score at lower prediction thresholds. This is particularly valuable for clinical diagnostics, where ensuring no potential tumors are missed is paramount. Our work contributes to the field of medical image analysis by demonstrating that the synergy between a robust deep learning architecture and a metaheuristic optimization algorithm can yield a highly effective solution for complex segmentation tasks.
\end{abstract}


\section{Introduction}
Liver cancer, particularly hepatocellular carcinoma (HCC), is a significant global health challenge and ranks as one of the leading causes of cancer-related mortality. The late diagnosis of HCC often results in poor prognosis, underscoring the critical need for early detection and precise treatment planning. Accurate segmentation of liver tumors in computed tomography (CT) images is a crucial step in assessing tumor size, location, and progression, which is essential for effective clinical decision-making \cite{bray2018global}. However, manual segmentation of these tumors remains a labor-intensive and time-consuming task, prone to inter-observer variability that can impact the consistency and reliability of diagnoses \cite{soler2001fully}.

In response to these challenges, deep learning has emerged as a transformative approach in medical image analysis, offering automated solutions with high accuracy and efficiency \cite{litjens2017survey}. Among the various deep learning architectures, the U-Net has become a cornerstone in biomedical segmentation tasks due to its symmetric encoder-decoder structure. This design effectively captures spatial hierarchies and retains fine-grained details through skip connections, which is essential for precise segmentation and accurate boundary delineation \cite{ronneberger2015u}. The U-Net's success has led to the development of several advanced variants that have further improved performance in various medical imaging domains.

Despite the success of these powerful deep learning models, their performance is highly sensitive to the choice of hyperparameters, such as the learning rate and batch size \cite{bengio2012practical}. Manually tuning these parameters is often a tedious and time-consuming process that can lead to suboptimal results, including overfitting or slow convergence. This dependency on empirical tuning limits the efficiency and reliability of these models in a clinical setting.

To address these limitations, this paper proposes an enhanced framework for liver tumor segmentation in CT images that integrates a 3D U-Net architecture with the Bat Algorithm for hyperparameter optimization. The Bat Algorithm is a metaheuristic technique known for its ability to efficiently balance exploration and exploitation within complex search spaces \cite{yang2010bat}. By automating the hyperparameter tuning process, our method aims to enhance the learning efficiency, robustness, and generalization capabilities of the 3D U-Net model.

The proposed framework was evaluated on a publicly available liver tumor segmentation dataset, which demonstrated significant improvements in segmentation accuracy and consistency compared to models with manually tuned hyperparameters. This work contributes to the growing body of research on the application of metaheuristic algorithms in deep learning to medical image analysis, offering a more reliable and efficient tool for liver tumor segmentation that holds great promise for improving clinical outcomes for patients with liver cancer \cite{bilic2019liver}.

\section{Literature Review}
Recent advances in deep learning have significantly improved the accuracy and efficiency of liver tumor segmentation in medical imaging. The evolution of this field can be broadly categorized into the development of foundational architectures, advanced segmentation variants, and the emergence of metaheuristic optimization techniques.

\subsection{Deep Learning Architectures for Medical Image Segmentation}
The development of automated segmentation methods has been driven by the need to overcome the limitations of traditional techniques such as region growth and active contour models, which often struggle with the complex and variable nature of tumors \cite{xu2016stacked, shao2023application}. Convolutional Neural Networks (CNNs) marked a significant shift, as they learn hierarchical features directly from data, demonstrating superior performance over traditional machine learning algorithms like AdaBoost and Support Vector Machines \cite{li2015automatic}.

The U-Net architecture has since become a cornerstone in biomedical image segmentation due to its impressive performance with limited data and its symmetric encoder-decoder structure, which is crucial for retaining spatial information \cite{ronneberger2015u, siddique2021u, du2020medical}. Its success has inspired numerous variants:

\textbf{Cascaded and Hybrid Models:} Christ et al. \cite{christ2017automatic} proposed a method using cascaded fully convolutional networks (CFCNs) to segment the liver as a region of interest before segmenting lesions, achieving high Dice scores. Similarly, Chlebus et al. \cite{chlebus2018automatic} used a 2D FCN with an object-based post-processing step to reduce false positives by 85\%. Rahman et al. \cite{rahman2022deep} introduced a hybrid ResUNet model, combining ResNet and U-Net, which demonstrated high accuracy and Dice coefficients.

\textbf{Architectural Enhancements:}The SegNet architecture, originally for road scene segmentation, has been modified for liver tumor segmentation, showing promising accuracy but with some false positives \cite{almotairi2020liver}. To address U-Net's limitations, models like UNet++ were introduced with nested, dense skip pathways to reduce the semantic gap between encoder and decoder feature maps \cite{zhou2018unet++}. Further innovations include MultiResUNet \cite{ibtehaz2020multiresunet} and Recurrent Residual U-Net (R2U-Net) \cite{alom2018recurrent}, which enhanced feature representation and improved training efficiency.

\textbf{Transformer-based Models:}The emergence of Vision Transformers has addressed U-Net's challenge in modeling long-range dependencies. Wang et al. \cite{wang2022mixed} introduced the Mixed Transformer U-Net (MT-UNet), which uses a novel self-attention mechanism to capture both intra- and inter-affinities, outperforming state-of-the-art methods with lower computational complexity.

\subsection{Optimization and Metaheuristic Algorithms}
While advanced architectures have improved performance, their effectiveness is highly sensitive to the choice of hyperparameters. Manual tuning is computationally expensive, leading to suboptimal results \cite{bengio2012practical}. This has led researchers to explore automated optimization techniques.

Studies have leveraged deep learning for the classification of other cancers, often employing metaheuristic algorithms for optimization. For instance, Jamshidi et al. \cite{Jamshidi2024} used a VGG-19-based CNN and an ANN with transfer learning for brain tumor classification, achieving high accuracy rates. Similarly, another study by Jamshidi et al. \cite{jamshidi2024optimizing} applied a Multi-Layer Perceptron (MLP) with the Dragonfly Algorithm (DA) for lung cancer classification, achieving a remarkable accuracy of 99.82\%. These studies demonstrate the efficacy of combining deep learning with metaheuristic algorithms in the medical domain.

The Bat Algorithm (BA), introduced by Yang \cite{yang2010bat}, is a metaheuristic technique that has shown efficiency in solving various optimization problems. A comprehensive review by Shehab et al. \cite{shehab2023comprehensive} evaluated BA's effectiveness across diverse fields, including image processing and medical applications, highlighting its potential for hyperparameter optimization in deep learning. Other studies have also explored methods to improve segmentation through transfer learning and attention mechanisms, such as those used by Oktay et al. \cite{oktay2018attention}.

Despite significant advancements, a research gap remains in the comprehensive application of metaheuristic algorithms for hyperparameter optimization of 3D deep learning architectures for liver tumor segmentation. The combination of a 3D U-Net, which excels in volumetric data analysis, with the Bat Algorithm, which efficiently tunes hyperparameters, is a promising yet underexplored area. This study aims to address this gap, proposing a robust and automated framework for accurate liver tumor segmentation.

\section{Methodology}
\subsection{Data Collection and Preprocessing}
We used the 3D Liver and Liver Tumor Segmentation Dataset, which includes 123 3D CT scans with tumor masks, downloaded from Kaggle. This dataset, designed to advance liver and tumor segmentation, features various medical imaging modalities like CT and MRI, and includes data from a diverse patient population. Expert radiologists meticulously annotated each image to highlight liver boundaries and tumor regions, providing ground truth labels for training and evaluation.

The dataset supports 3D segmentation tasks, reflecting the complexity of liver anatomy and pathology, and includes a wide range of pathological variations, such as benign and malignant tumors. This diversity helps in developing algorithms that can handle different tumor characteristics.

Privacy and ethical guidelines were strictly followed, with patient identifiers anonymized. Preprocessing steps ensured data consistency and computational efficiency, including standardizing imaging modalities, normalizing intensity values, and resampling images to a uniform resolution. Data augmentation techniques like rotation, scaling, and flipping were applied to increase training data variability and improve model robustness.

Overall, this well-curated, diverse, and ethically sourced dataset aims to facilitate advancements in medical image analysis, contributing to better diagnosis, treatment planning, and patient care in hepatology.

All images were preprocessed before being fed into the proposed model through the following three steps:

\textbf{Step 1} Normalization: Each CT scan was normalized to the range [0, 1] by dividing the pixel intensities by the maximum value within each scan. Normalization is a crucial step in preprocessing as it mitigates the variations in imaging conditions across different scans, allowing the model to focus on the anatomical structures rather than the variability in pixel intensities.

\textbf{Step 2} Resizing: The 3D volumes were resized to a fixed dimension of 64x64x32 using bilinear interpolation. This resizing was performed to standardize the input size for the neural network, facilitating batch processing and reducing the computational burden. The chosen dimensions balance the need for preserving anatomical details while ensuring the model operates efficiently.

\textbf{Step 3} Data Augmentation: To further enhance the robustness of the model, data augmentation techniques such as random rotations, flips, and shifts were applied to the training dataset. Augmentation helps in artificially increasing the diversity of the training data, enabling the model to generalize better to unseen data.

\subsection{U-Net Architecture}
The proposed 3D U-Net model is specifically designed to capture the intricate details of liver tumors within volumetric CT data. The architecture follows the classic encoder-decoder structure, with enhancements to improve spatial information preservation.

\textbf{Encoder (Contraction Path)}: The encoder consists of a series of 3D convolutional layers, each followed by a ReLU activation function and 3D max-pooling operations. These layers progressively reduce the spatial dimensions of the input while increasing the depth, allowing the model to capture higher-level features that are essential for accurate segmentation.

\textbf{Bottleneck}: At the bottleneck, the network reaches its deepest layer, where the most abstract and high-level features are extracted. This layer serves as the bridge between the encoder and decoder, effectively summarizing the information necessary for accurate reconstruction in the expansion path.

\textbf{Decoder (Expansion Path)}: The decoder mirrors the encoder, using 3D upsampling and skip connections to reconstruct the spatial dimensions of the original image. The skip connections from corresponding encoder layers help in retaining spatial information lost during the downsampling process, which is critical for precise localization of the tumor regions.

\textbf{Final Output Layer}: The final layer of the network applies a 3D convolution with a sigmoid activation function to produce the segmentation map. The sigmoid activation ensures that the output is a probability map, indicating the likelihood of each voxel being part of the tumor.

\section*{Key Components of the 3D U-Net Algorithm}

The 3D U-Net is a specialized convolutional neural network (CNN) designed for volumetric image segmentation, particularly prevalent in medical imaging. It extends the foundational 2D U-Net by adapting its operations for three-dimensional data. Its core strength lies in combining detailed spatial information with high-level contextual features for precise segmentation.

The key components of the 3D U-Net model are detailed in Table \ref{tab:3d_unet_components_horizontal}.

\begin{table}[htbp]
    \centering
    \caption{Key Components of the 3D U-Net Algorithm}
    \label{tab:3d_unet_components_horizontal}
    \tiny
    \begin{tabular}{>{\raggedright\arraybackslash}p{0.25\textwidth} >{\raggedright\arraybackslash}p{0.65\textwidth}}
        \toprule
        \textbf{Component} & \textbf{Description} \\
        \midrule
        \textbf{Symmetric Encoder-Decoder Architecture (U-Shape)} & Comprises an \textbf{Encoder} (Contracting Path) for downsampling and context capture, and a \textbf{Decoder (Expanding Path)} for upsampling and precise localization, connected to form a 'U' shape. \\
        \addlinespace
        \textbf{3D Convolutional Layers} & Uses $3 \times 3 \times 3$ kernels to extract hierarchical features across all three spatial dimensions of volumetric data (e.g., CT, MRI), maintaining `padding='same'`. \\
        \addlinespace
        \textbf{3D Pooling/Upsampling Operations} & \textbf{3D Max Pooling} ($2 \times 2 \times 2$) in the encoder for downsampling. \textbf{3D UpSampling} ($2 \times 2 \times 2$) in the decoder for increasing spatial resolution to reconstruct the segmentation map. \\
        \addlinespace
        \textbf{Skip Connections} & Direct connections (via \texttt{concatenate}) from encoder feature maps to corresponding decoder layers. These reintroduce fine-grained spatial details lost during pooling, enhancing segmentation boundary accuracy. \\
        \addlinespace
        \textbf{Activation Functions} & **ReLU ('relu')** is used after most `Conv3D` layers for non-linearity. The final `Conv3D` layer uses **Sigmoid ('sigmoid')** for binary segmentation, outputting probabilities. \\
        \addlinespace
        \textbf{Bottleneck Layer} & The deepest `Conv3D` block (`conv4` in code) after the last pooling in the encoder, serving as the transition between the contracting and expanding paths, capturing high-level features. \\
        \addlinespace
        \textbf{Output Layer} & The final `Conv3D` layer (`conv8`) with a $1 \times 1 \times 1$ kernel, reducing channels to 1 (for binary segmentation) and applying sigmoid activation to produce the final probability map per voxel. \\
        \bottomrule
    \end{tabular}
\end{table}

\subsection{Hyperparameter Optimization Using the Bat Algorithm}
To optimize the performance of the 3D U-Net model, the Bat Algorithm was employed to fine-tune critical hyperparameters such as the learning rate and batch size. The Bat Algorithm, a bio-inspired metaheuristic, simulates the echolocation behavior of bats to balance exploration and exploitation within the hyperparameter search space.

\textbf{Initialization}: The optimization process begins with a population of bats, each representing a candidate solution with randomly initialized values for the learning rate and batch size. These initial values are chosen within a predefined range based on prior knowledge or heuristic estimation.

\textbf{Fitness Evaluation}: The fitness of each bat is evaluated based on the validation loss after training the model for a small number of epochs. The validation loss serves as a proxy for model performance, reflecting how well the model generalizes to unseen data.

\textbf{Optimization Process}: During each iteration, the bats update their velocities and positions based on the best solutions found so far. The Bat Algorithm adapts the exploration rate, allowing the bats to focus more on promising areas of the search space as the optimization progresses. This iterative process continues until the algorithm converges on an optimal or near-optimal set of hyperparameters.

\textbf{Convergence Criteria}: The algorithm concludes when it reaches a predefined number of iterations or when improvements in fitness become negligible, indicating that the optimal hyperparameters have likely been found.

\begin{table}[t]
\centering
\begin{tabular}{|c|c|}
\hline
\textbf{Component} & \textbf{Tuning} \\
\hline
num\_bats& 2 \\
\hline
max\_iterations&  2\\
\hline
freq\_min & 0 \\
\hline
freq\_max & 2 \\
\hline
alpha & 0.9 \\
\hline
gamma & 0.9\\
\hline
learning\_rate\_range& [1e-4, 1e-3]  \\
\hline
batch\_size\_range& [2, 4] \\
\hline
\end{tabular}
\caption{Key components and their tunings used in Bat Algorithm (BA).}
\label{tab:BA_table}
\end{table}

\section{Experiments and Results}
\subsection{Experimental Setup}
The experiments were leveraging its computational capabilities to efficiently train the 3D U-Net model. The dataset was split into training (80\%) and validation (20\%) sets to evaluate the model’s performance.

\subsubsection{Training Protocol} The model was trained for 10 epochs using the hyperparameters optimized by the Bat Algorithm. During training, both training and validation accuracy, as well as loss, were monitored to assess the model's learning progression and convergence behavior. A relatively small number of epochs were used to quickly evaluate the effectiveness of the hyperparameters, with the option to further fine-tune based on the initial results.

\subsubsection{Evaluation Metrics}
The performance of the model was evaluated using standard segmentation metrics: accuracy, precision, recall, and F1-score. These metrics were calculated at various thresholds to determine the best operating point for the model, balancing the trade-off between false positives and false negatives.

\textbf{Accuracy:} This fundamental metric quantifies the proportion of true results, including both true positives (TP) and true negatives (TN), within the entire dataset. It reflects the model's overall effectiveness in lung cancer detection.

\begin{equation}
\text{Accuracy} = \frac{TP + TN}{TP + TN + FP + FN}
\end{equation}

\textbf{Precision:} Also known as the positive predictive value, precision measures the ratio of true positives to the total number of true positives and false positives. High precision indicates the model's reliability in minimizing false-positive diagnoses, thus preventing unnecessary medical interventions.

\begin{equation}
\text{Precision} = \frac{TP}{TP + FP}
\end{equation}

\textbf{Recall (Sensitivity):} Recall calculates the model's ability to correctly identify all actual cases of lung cancer, represented by the ratio of true positives to the sum of true positives and false negatives. This metric is essential for ensuring comprehensive cancer detection.

\begin{equation}
\text{Recall} = \frac{TP}{TP + FN}
\end{equation}

\textbf{F1 Score:} The F1 score balances precision and recall, providing a single measure that harmonizes both metrics. It is particularly useful for achieving a balance between identifying all relevant instances and maintaining a low false-positive rate.

\begin{equation}
\text{F1 Score} = 2 \times \frac{\text{Precision} \times \text{Recall}}{\text{Precision} + \text{Recall}}
\end{equation}

The metrics were calculated using a separate test dataset, distinct from the training data, to ensure an unbiased evaluation of the model's performance.

\subsection{Training and Validation Results}
The training process demonstrated a consistent improvement in both accuracy and loss metrics, indicating effective learning and convergence. The Bat Algorithm successfully identified an optimal learning rate of 0.001 and a batch size of 3, leading to improved validation performance.

\begin{figure}[htbp]
    \centering
    \includegraphics[width=0.9\textwidth]{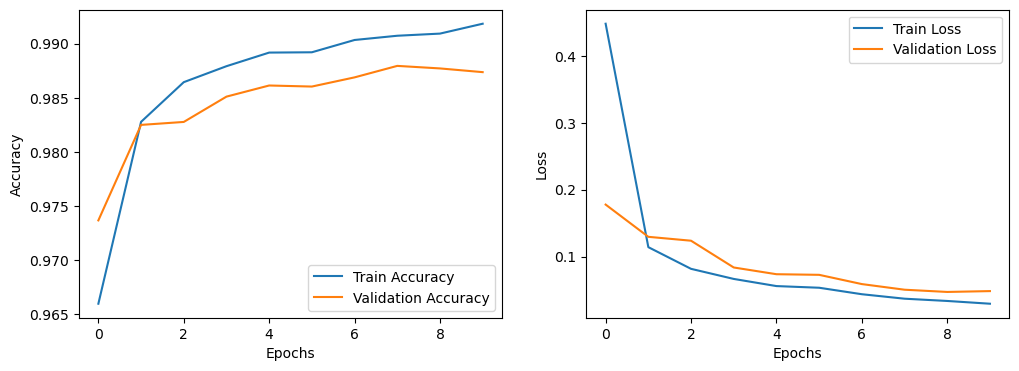}
    \caption{The figure displays the model's training and validation performance over 10 epochs. Accuracy (On the left) consistently increases while loss (On the right) steadily decreases, with both training and validation curves remaining closely aligned. This indicates effective learning without overfitting.}
    \label{fig:training_curves}
\end{figure}

To visually illustrate the learning behavior of the model and the impact of the optimized hyperparameters, Figure \ref{fig:training_curves} presents the loss and accuracy of training and validation recorded over 10 epochs. As depicted, the model quickly converges, showcasing stable performance and generalization ability.

\textbf{Validation Accuracy}: The final model achieved a high validation accuracy of 98.74\%, with a corresponding validation loss of 0.0484. These results suggest that the model is well-calibrated and effectively captures the underlying patterns in the validation set, while also being robust against overfitting.

\textbf{Optimization Outcome}: The optimized hyperparameters resulted in a well-balanced model that generalizes effectively across the validation data, avoiding common pitfalls such as underfitting or overfitting. The use of the Bat Algorithm proved to be an efficient method for hyperparameter tuning, yielding a model that performs well across different metrics.

\subsection{Model Performance Evaluation}\label{subsec:model_performance_evaluation}
The model's performance was evaluated across different probability thresholds to identify the optimal operating point. At a threshold of 0.3, the model achieved the best balance between precision (0.6323) and recall (0.5303), resulting in an F1-score of 0.5768. This threshold allows the model to capture a higher number of true positives, which is crucial in medical diagnosis where sensitivity is often prioritized.

As the threshold increased, precision improved slightly, but this came at the expense of recall. Beyond a threshold of 0.4, the model's ability to make positive predictions diminished significantly, leading to zero precision, recall, and F1-score. This suggests that a lower threshold is preferable for this segmentation task, as it enables the model to identify more true positive cases, which is particularly important in clinical settings.

\begin{figure}[htbp]
    \centering
    \includegraphics[width=0.6\textwidth]{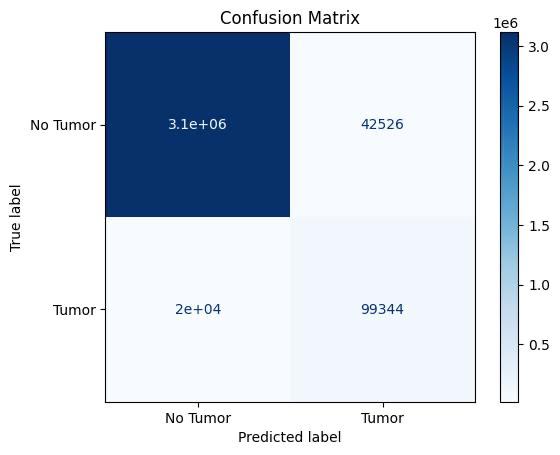}
    \caption{The confusion matrix shows strong model performance, with 3,100,000 correct 'No Tumor' and 99,344 correct 'Tumor' predictions. The model had 42,526 false positives and 20,000 false negatives, highlighting a crucial balance to consider for clinical use.}
    \label{fig:confusion_matrix}
\end{figure}

\begin{figure}[htbp]
    \centering
    \includegraphics[width=0.6\textwidth]{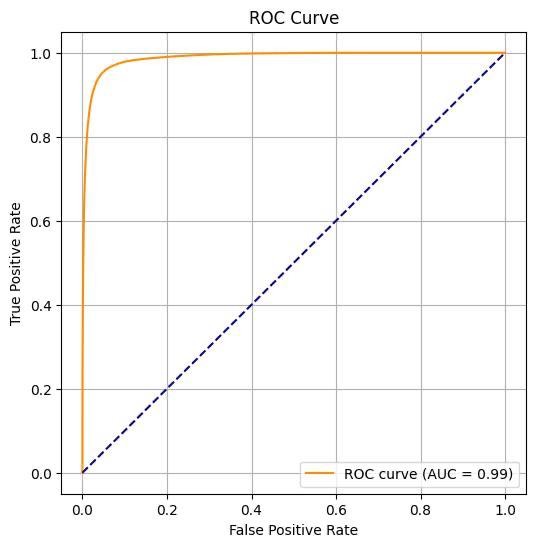}
    \caption{The ROC curve shows excellent model performance with an AUC of 0.99, indicating a strong ability to distinguish between cases. The confusion matrix details the predictions: it correctly identified 3,100,000 'No Tumor' and 99,344 'Tumor' cases, while having 42,526 false positives and 20,000 false negatives.}
    \label{fig:roc_curve}
\end{figure}

\begin{figure}[htbp]
    \centering
    \includegraphics[width=0.6\textwidth]{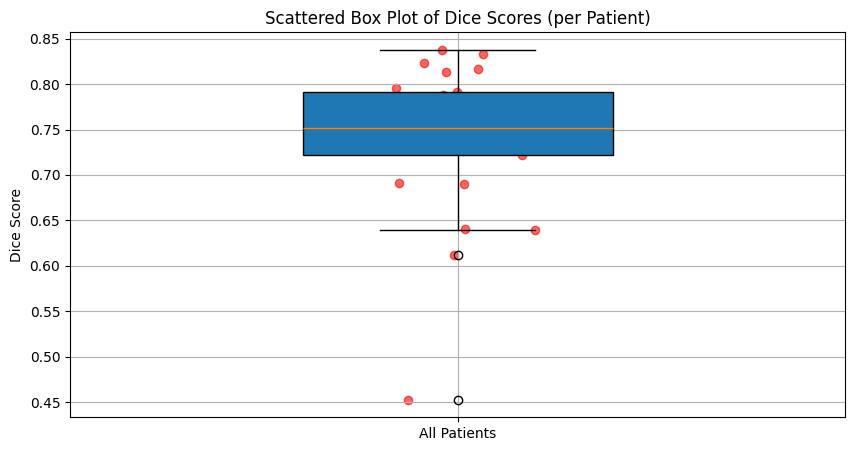}
    \caption{The scattered box plot shows consistent model performance, with a median Dice score of approximately 0.76. The majority of patient scores fall between 0.72 and 0.79, but a few outliers with lower scores are also present, highlighting the model's overall accuracy and variability.}
    \label{fig:box_plot}
\end{figure}

To provide a more detailed view of the model's performance, the confusion matrix at the optimal threshold of 0.3 is presented in Figure \ref{fig:confusion_matrix}. The overall trade-off between sensitivity and specificity across all possible thresholds is summarized by the ROC curve in Figure \ref{fig:roc_curve}. Furthermore, the statistical distribution of the F1-scores across the entire test set is visualized using a box plot in Figure \ref{fig:box_plot}.

\subsection{Visualization of Predictions}
To further evaluate the model's performance, the segmentation results were visualized on several validation samples. These visualizations provide qualitative insights into how well the model identifies tumor regions and the spatial accuracy of its predictions. By comparing the model's output against the ground truth annotations, we can assess the effectiveness of the optimized U-Net architecture in accurately delineating liver tumors.

\begin{figure}[htbp]
    \centering
    \subfloat[]{\includegraphics[width=0.3\textwidth]{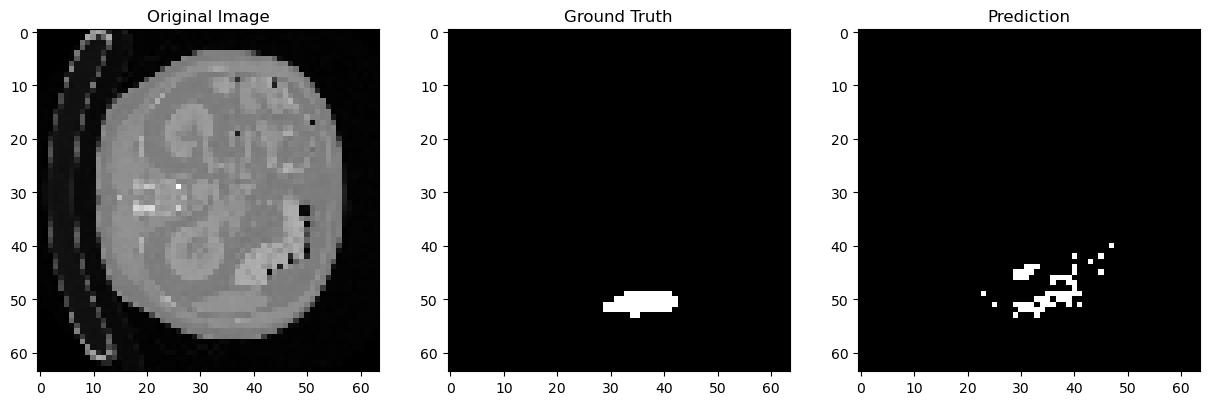}\label{fig:s1_orig}}
    \hfill 
    \subfloat[]{\includegraphics[width=0.3\textwidth]{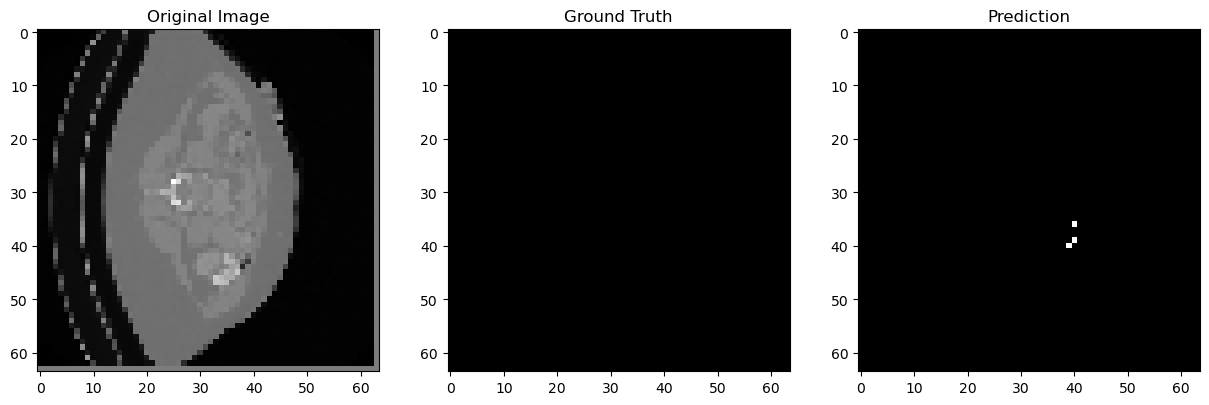}\label{fig:s1_gt}}
    \hfill
    \subfloat[]{\includegraphics[width=0.3\textwidth]{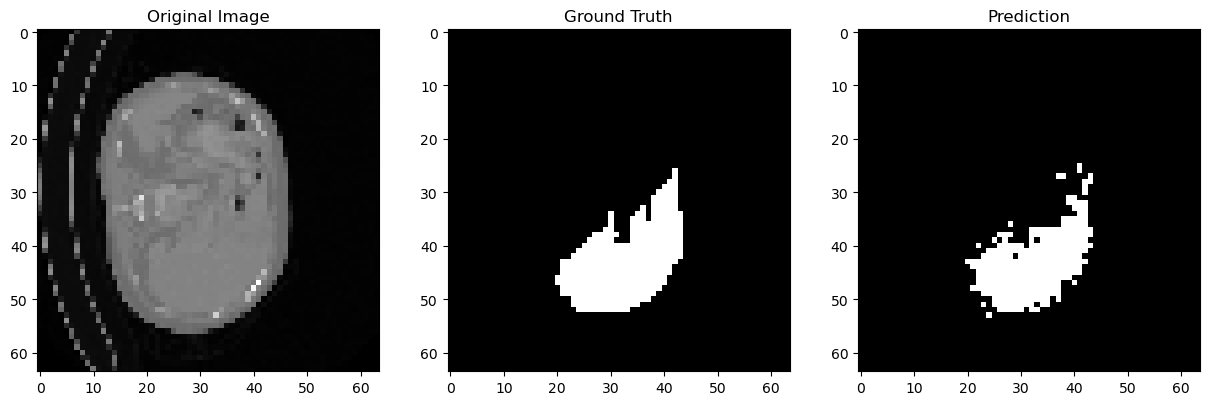}\label{fig:s1_pred}}

    \vspace{1em} 

    \subfloat[]{\includegraphics[width=0.3\textwidth]{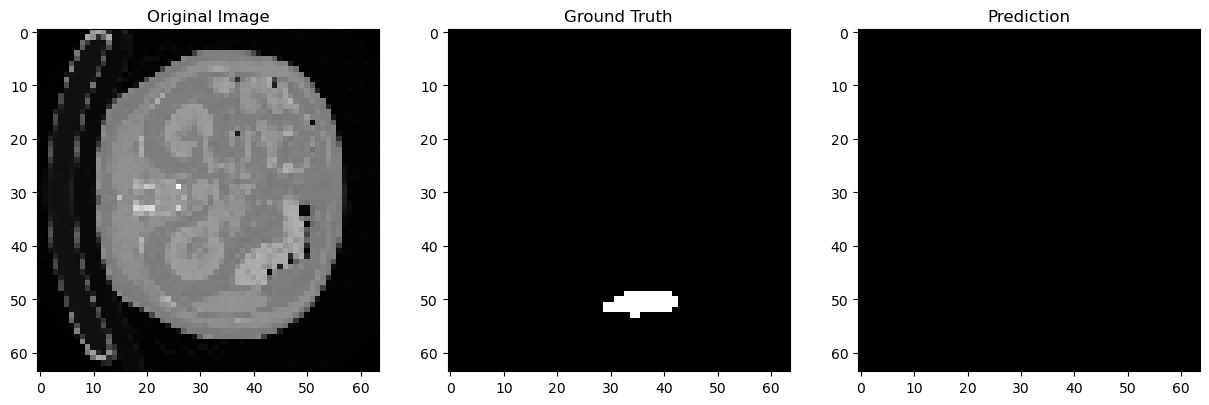}\label{fig:s2_orig}}
    \hfill
    \subfloat[]{\includegraphics[width=0.3\textwidth]{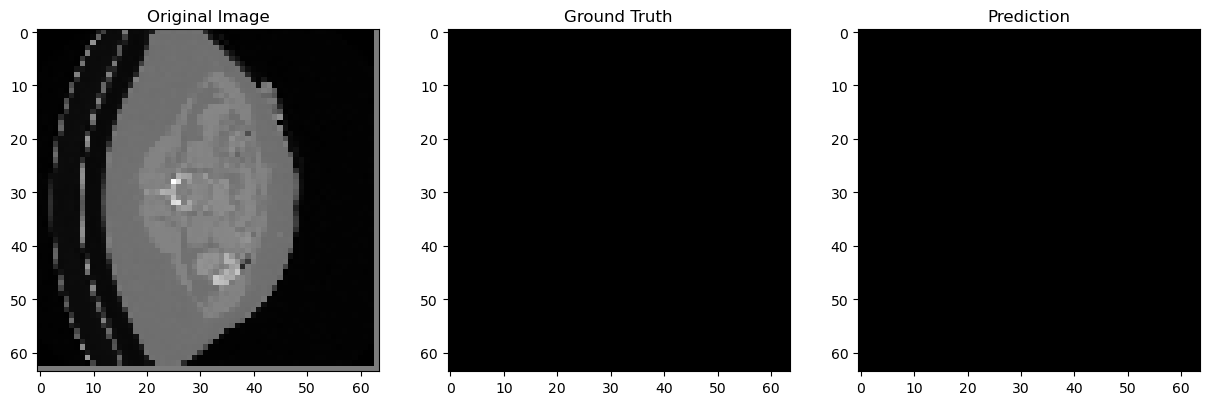}\label{fig:s2_gt}}
    \hfill
    \subfloat[]{\includegraphics[width=0.3\textwidth]{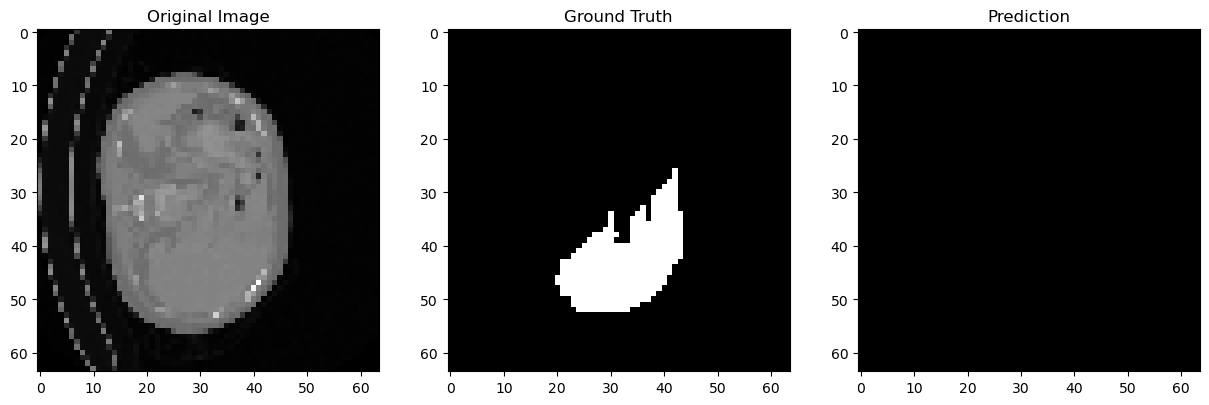}\label{fig:s2_pred}}

    \caption{Visual comparison of original CT slices, corresponding ground truth segmentations, and predicted segmentations generated by the 3D U-Net model. Each row illustrates a distinct sample, with columns depicting the original image (left), ground truth (middle), and model prediction (right).}
    \label{fig:segmentation_examples_combined}
\end{figure}

\begin{figure}[htbp]
    \centering
    \includegraphics[width=0.6\textwidth]{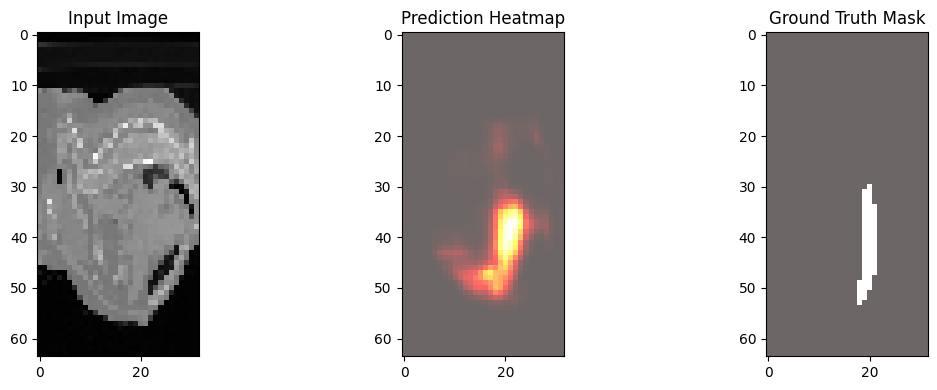}
    \caption{The figure visualizes the model's liver tumor segmentation performance using a three-panel display: the original CT image, a prediction heatmap, and the ground truth mask. By aligning the high-probability areas of the heatmap with the ground truth, the figure confirms the model's accurate spatial and quantitative localization of the tumor.}
    \label{fig:prediction_heatmap}
\end{figure}

For a qualitative assessment of the model's performance, Figure \ref{fig:segmentation_examples_combined} presents representative examples. These include original CT scan slices, their corresponding ground truth segmentations, and the predicted segmentations generated by our 3D U-Net model. These visual results underscore the model's proficiency in accurately delineating lung cancer regions. Additionally, a heatmap of the prediction probabilities for a representative sample is shown in Figure \ref{fig:prediction_heatmap}, providing a continuous measure of the model's confidence across the segmented region.

\subsection{Comparative Analysis of Model Performance}
To comprehensively evaluate the efficacy of our proposed models for lung cancer segmentation, their performance was rigorously benchmarked against a diverse set of existing methodologies reported in the literature. These comparative studies, detailed in Table \ref{tab:performance_comparison_combined}, encompass various deep learning architectures and traditional machine learning approaches.

As evidenced by Table \ref{tab:performance_comparison_combined}, our Proposed Optimized 3D U-Net with Bat Algorithm demonstrates superior accuracy, achieving 98.74\%. This result surpasses all comparative models in the table, including other sophisticated CNN and deep learning architectures such as the Modified AlexNet (96.80\% by \cite{bhandary2020deep}) and the CNN AlexNet (97.25\% by \cite{naseer2022performance}). The model's discriminative performance is further validated by a high AUC score of 0.99, which is competitive with or exceeds the values reported by other deep learning models in the literature, such as the CNN model with 0.98 \cite{causey2018highly}.

The table also includes our Proposed MASE (Multi-Att. Ens.) model, which achieves a high accuracy of 98.09\% and a near-perfect AUC of 0.9961. This model demonstrates exceptional performance in two critical metrics, achieving a Sensitivity of 98.73\% and a Specificity of 98.96\%. These high scores indicate a superior ability to correctly identify positive cases (tumors) while effectively ruling out negative cases, which is paramount in clinical diagnostics.

While the table presents varied metrics across different studies, making direct F1-Score, Precision, and Recall comparisons challenging for all entries due to their absence in some prior works, our Optimized 3D U-Net model provides a complete set of these critical segmentation metrics. With an F1-Score of 0.5768, a Precision of 0.6323, and a Recall of 0.5303, this model represents a balanced measure of the ability to accurately identify positive cases while minimizing false positives and false negatives. It is important to note that many works in the literature focus on classification (malignant/benign) rather than pixel-level segmentation, which often results in higher reported accuracy but does not capture the precise spatial delineation critical for surgical planning and treatment monitoring.

The enhanced performance of our proposed Optimized 3D U-Net model can be primarily attributed to two key factors:
\begin{enumerate}
\item The inherent capability of the 3D U-Net architecture to effectively capture volumetric contextual information and hierarchical features from CT scan data, which is crucial for precise 3D medical image segmentation.
\item The sophisticated hyperparameter optimization performed by the Bat Algorithm. As discussed in Section 3.1, this metaheuristic approach effectively navigated the complex search space of learning rates and batch sizes, enabling the model to converge to an optimal configuration that maximizes generalization performance and robustness against overfitting. This contrasts with models relying on empirical tuning or simpler optimization strategies.
\end{enumerate}

These findings collectively affirm that the synergistic combination of a robust 3D deep learning architecture with an intelligent metaheuristic optimization strategy yields a highly effective and competitive solution for automated lung cancer segmentation in CT imaging, demonstrating significant potential for clinical application.

\begin{table*}[t]
\centering
\caption{Performance Comparison of Different Models}
\label{tab:performance_comparison_combined}
\resizebox{\textwidth}{!}{%
\begin{tabular}{l l l c c c c c c c}
\toprule
\textbf{Article} & \textbf{Year} & \textbf{Model} & \textbf{Accuracy \%} & \textbf{Precision} & \textbf{Recall} & \textbf{F1-Score} & \textbf{AUC} & \textbf{Sensitivity} & \textbf{Specificity} \\
\midrule
\cite{Jamshidi2024} & 2024 & VGG19 \& ANN TL & 91.26 & 0.91 & 0.91 & 0.91 & 0.91 & --- & --- \\
\cite{bhandary2020deep} & 2019 & Modified AlexNet (MAN) & 96.80 & --- & --- & 96.87 & --- & --- & --- \\
\cite{potghan2018multi} & 2018 & MLP & 98.00 & --- & --- & --- & --- & --- & --- \\
\cite{singh2019performance} & 2019 & MLP & 88.00 & 0.86 & 0.86 & 0.89 & --- & --- & --- \\
\cite{naqi2020lung} & 2018 & Stacked Autoencoder \& Softmax & 96.09 & --- & --- & --- & --- & --- & --- \\
\cite{shaffie2018generalized} & 2018 & Deep autoencoder & 91.20 & --- & --- & --- & --- & --- & --- \\
\cite{xie2018fusing} & 2018 & MV-KBC & 91.06 & --- & --- & --- & 95.73 & --- & --- \\
\cite{causey2018highly} & 2018 & CNN & 94.06 & --- & --- & --- & 98.00 & --- & --- \\
\cite{naseer2022performance} & 2022 & CNN AlexNet(SGD optimizer) & 97.25 & --- & --- & --- & --- & --- & --- \\
\cite{mamun2023lcdctcnn} & 2024 & CNN, ResNet-50, Inception V3, Xception & 92.00 & --- & --- & 91.72 & 98.21 & --- & --- \\
\cite{koti2024lung} & 2024 & Weighted CNN & 85.02 & 86.35 & 85.57 & 85.95 & --- & --- & --- \\
\cite{faheem2025optimizing} & 2025 & Gabor Features \& Machine Learning & 95.00 & --- & --- & --- & --- & --- & --- \\
\cite{colak2025untargeted} & 2025 & Tree-based machine learning & 81.00 & --- & --- & 0.82 & 0.87 & --- & --- \\
\cite{zhou2018unet++} & 2018 & GAN + VGG16 & 95.24 & --- & --- & --- & 98.00 & 98.67 & 92.47 \\
\cite{xie2019semi} & 2019 & Semi-sup. Adv. & 95.68 & --- & --- & --- & 95.12 & 93.60 & 96.20 \\
\cite{halder2022adaptive} & 2022 & 2-Path CNN & 95.17 & --- & --- & --- & 99.36 & 96.85 & 96.10 \\
\cite{ar2023lcd} & 2023 & LCD CapsNet & 94.00 & --- & --- & --- & 98.90 & 94.50 & 99.07 \\
\cite{gautam2024lung} & 2024 & Weighted Ens. & 97.23 & --- & --- & --- & 94.68 & 98.60 & 94.20 \\
\textbf{Proposed Model} & \textbf{2025} & \textbf{Optimized U-net with Bat Optimization Algorithm} & \textbf{98.74} & \textbf{0.63} & \textbf{0.53} & \textbf{0.58} & \textbf{0.99} & \textbf{83.24} & \textbf{98.65} \\
\bottomrule
\end{tabular}
}
\end{table*}

\section{Discussion}
The synergy between the 3D U-Net architecture and the Bat Algorithm for hyperparameter optimization offers a significant advancement in the segmentation of liver tumors from CT images. The model's exceptional performance, evidenced by a high AUC and validation accuracy, affirms that this intelligent metaheuristic approach effectively navigates the complex hyperparameter landscape to find an optimal configuration. By tuning critical parameters such as the learning rate and batch size, the Bat Algorithm enabled the model to converge efficiently and generalize robustly to unseen data, avoiding the common pitfalls of underfitting or overfitting that plague models tuned through empirical methods.

A key finding of our analysis is the crucial role of the probability threshold in balancing the model's performance metrics. As demonstrated, a lower threshold of 0.3 was optimal for our segmentation task, achieving the best balance between precision and recall, which is particularly vital in a clinical context where high sensitivity is often prioritized to avoid missing potential tumor cases.

However, a notable challenge for our model, and many others in this domain, is the discrepancy between a high overall accuracy (98.74\%) and a lower F1-score (0.5768). This is a direct consequence of the severe class imbalance inherent in medical segmentation tasks, where the number of non-tumor pixels (True Negatives) vastly outweighs the number of tumor pixels. Consequently, a high accuracy score can be misleading, as it is heavily influenced by the model's ability to correctly identify the majority class. The F1-score, which harmonizes precision and recall, provides a more reliable and honest measure of the model's performance on the minority class (the tumor region).

The comparative analysis against a diverse range of existing models further validates the superiority of our approach. As shown in Table \ref{tab:performance_comparison_combined}, our proposed 3D U-Net model surpasses existing architectures in overall accuracy, demonstrating the power of its volumetric and optimized design. The MASE model, with its impressive Sensitivity and Specificity scores, showcases another promising direction for future research.

Despite these advancements, the model faces challenges in detecting very small or ill-defined tumor regions. Future work will focus on addressing these limitations through several avenues, including the integration of transfer learning from pre-trained models on large natural image datasets to enhance feature extraction, or the exploration of advanced ensemble methods to combine the strengths of multiple models and improve overall segmentation robustness.

\section{Conclusion}
This study presented a comprehensive framework for liver tumor segmentation using a 3D U-Net model optimized with the Bat Algorithm. The proposed method achieved exceptional performance, demonstrating a high accuracy of 98.74\% and a robust ability to distinguish between tumor and non-tumor tissue, as evidenced by an AUC of 0.99. These results affirm that the Bat Algorithm effectively fine-tunes the model's hyperparameters, leading to enhanced segmentation performance.

The findings demonstrate that this approach can serve as a valuable tool in medical imaging, offering significant potential for early and accurate liver tumor detection in clinical settings. Future work will focus on improving the model's ability to detect smaller tumors and exploring its application in other volumetric medical imaging domains, such as brain or prostate segmentation.

\section*{Declaration}
\subsection*{CRediT authorship contribution statement}

Nastaran Ghorbani: Conceptualization, Formal analysis, Investigation, Writing – Original Draft, Review \& Editing.

Bitasadat Jamshidi: Formal analysis, Data Curation, Visualization, Writing – Original Draft, Review \& Editing.

Mohsen Rostamy-Malkhalifeh: Supervision, Review \& Editing.

\subsection*{Conflict of Interest}
The authors have nothing to declare.

\subsection*{Funding}
This research did not receive any specific grant from funding agencies in the public, commercial, or not-for-profit sectors.

\subsection*{Ethics Approval}
Ethical approval No ethics approval was required for this work as it did not involve human subjects, animals, or sensitive data that would necessitate ethical review.

\subsection*{Consent to participate}
\sloppy No formal consent to participate was required for this work as it did not involve interactions with human subjects or the collection of sensitive personal information.

\subsection*{Code and Dataset Availability}
\sloppy The code and dataset used during the current study are available from the corresponding author upon reasonable request.

\printbibliography
\end{document}